\documentclass[journal,twoside,web]{ieeecolor}
\usepackage{tmi}
\usepackage{cite}
\usepackage{amsmath,amssymb,amsfonts}
\usepackage{algorithmic}
\usepackage{graphicx}
\usepackage{textcomp}

\usepackage{booktabs}
\usepackage{multirow}
\usepackage{subcaption}
\usepackage{amsmath}
\usepackage{amssymb}
\usepackage{url}
\usepackage{caption}
\usepackage{wrapfig}
\usepackage{cite}
\usepackage{threeparttable}
\usepackage{adjustbox}
\usepackage{color}

\def\BibTeX{{\rm B\kern-.05em{\sc i\kern-.025em b}\kern-.08em
    T\kern-.1667em\lower.7ex\hbox{E}\kern-.125emX}}
\begin{document}
\title{Deep Omni-supervised Learning for Rib Fracture Detection from Chest Radiology Images}
\author{Zhizhong~Chai,  Luyang~Luo, \IEEEmembership{Member, IEEE}, Huangjing~Lin,  Pheng-Ann~Heng, \IEEEmembership{Senior~Member, IEEE}, and~Hao~Chen, \IEEEmembership{Senior~Member, IEEE}
\thanks{This work was supported by the Research Grants Council of the Hong Kong Special Administrative Region, China (Project Reference Number: T45-401/22-N), Hong Kong Innovation and Technology Fund (Project No. ITS/028/21FP), Shenzhen Science and Technology Innovation Committee Funding (Project No. SGDX20210823103201011), and the Project of Hetao Shenzhen-Hong Kong Science and Technology Innovation Cooperation Zone (HZQB-KCZYB-2020083).}
\thanks{Zhizhong~Chai and Luyang~Luo contributed equally.}
\thanks{Zhizhong~Chai and Huangjing~Lin are with Imsight AI Research Lab, Shenzhen, China (e-mail: chaizhizhong, linhuangjing@imsightmed.com)}
\thanks{Luyang~Luo is with the Department of Computer Science and Engineering, The Hong Kong University of Science and Technology, Hong Kong, China (e-mail: cseluyang@ust.hk).}
\thanks{Hao Chen is with the Department of Computer Science and Engineering and Department of Chemical and Biological Engineering, Hong Kong University of Science and Technology, Hong Kong, China. 
Hao Chen is also affiliated with HKUST Shenzhen-Hong Kong Collaborative Innovation Research Institute, Futian, Shenzhen, China. (e-mail: jhc@cse.ust.hk)}
\thanks{ Pheng-Ann Heng is with the Department of Computer Science and Engineering, The Chinese University of Hong Kong, Hong Kong, China (e-mail: pheng@cse.cuhk.edu.hk).}
\thanks{Hao Chen is the corresponding author.}}

\maketitle

\begin{abstract}
Deep learning (DL)-based rib fracture detection has shown promise of playing an important role in preventing mortality and improving patient outcome.
Normally, developing DL-based object detection models requires a huge amount of bounding box annotation.
However, annotating medical data is time-consuming and expertise-demanding, making obtaining a large amount of fine-grained annotations extremely infeasible.
This poses a pressing need {for} developing label-efficient detection models to alleviate radiologists' labeling burden.
To tackle this challenge, the literature on object detection has witnessed an increase of weakly-supervised and semi-supervised approaches, yet still lacks a unified framework that leverages various forms of fully-labeled, weakly-labeled, and unlabeled data.
In this paper, we present a novel omni-supervised object detection network, ORF-Netv2, to leverage as much available supervision as possible.
Specifically, a multi-branch omni-supervised detection head is introduced with each branch trained with a specific type of supervision. 
A co-training-based dynamic label assignment strategy is then proposed to enable flexible and robust learning from the weakly-labeled and unlabeled data.
Extensive evaluation was conducted for the proposed framework with three rib fracture datasets on both chest CT and X-ray.
By leveraging all forms of supervision, ORF-Netv2 achieves mAPs of 34.7, 44.7, and 19.4 on the three datasets, respectively, surpassing the baseline detector which uses only box annotations by mAP gains of 3.8, 4.8, and 5.0, respectively.
Furthermore, ORF-Netv2 consistently outperforms other competitive label-efficient methods over various scenarios, showing a promising framework for label-efficient fracture detection.
The code is available at: \url{https://github.com/zhizhongchai/ORF-Net}.
\end{abstract}


\section{Introduction}
\label{sec:introduction}
\IEEEPARstart{R}{ib} fracture is the most common form of blunt thoracic injury \cite{talbot2017traumatic}. Many studies highlighted that high morbidity and mortality can be associated with even a single rib fracture and increase with the number of rib fractures \cite{talbot2017traumatic, haines2018rib}. 
In addition, the diagnosis of rib fractures helps determine the severity of the trauma.
Therefore, accurate recognition and location of rib fractures are of significant clinical value for preventing mortality and improving patient outcome.

\begin{figure}[t]
\centering
\begin{subfigure}{0.47\columnwidth}
  \centering
  \includegraphics[width=\linewidth]{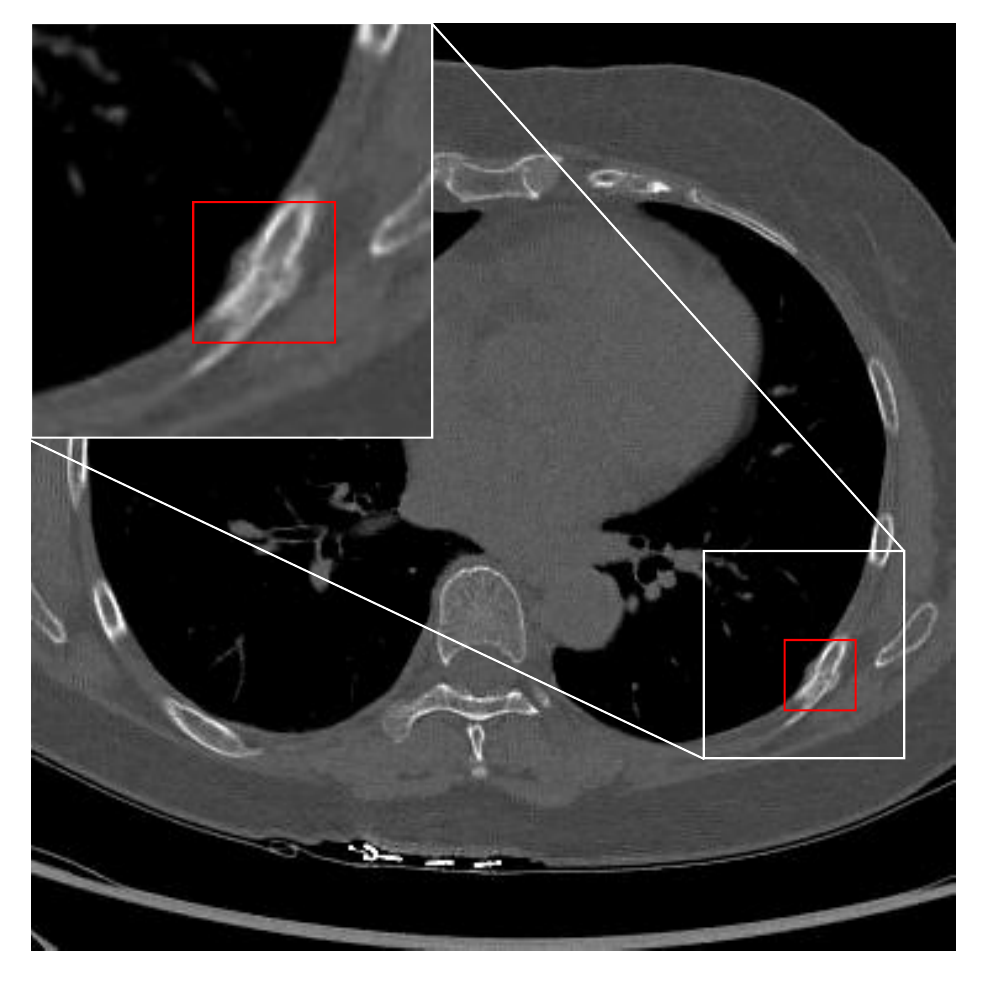}
  \caption{Box-labeled CT}
  \label{fig:a}
\end{subfigure}
\begin{subfigure}{0.47\columnwidth}
  \centering
  \includegraphics[width=\linewidth]{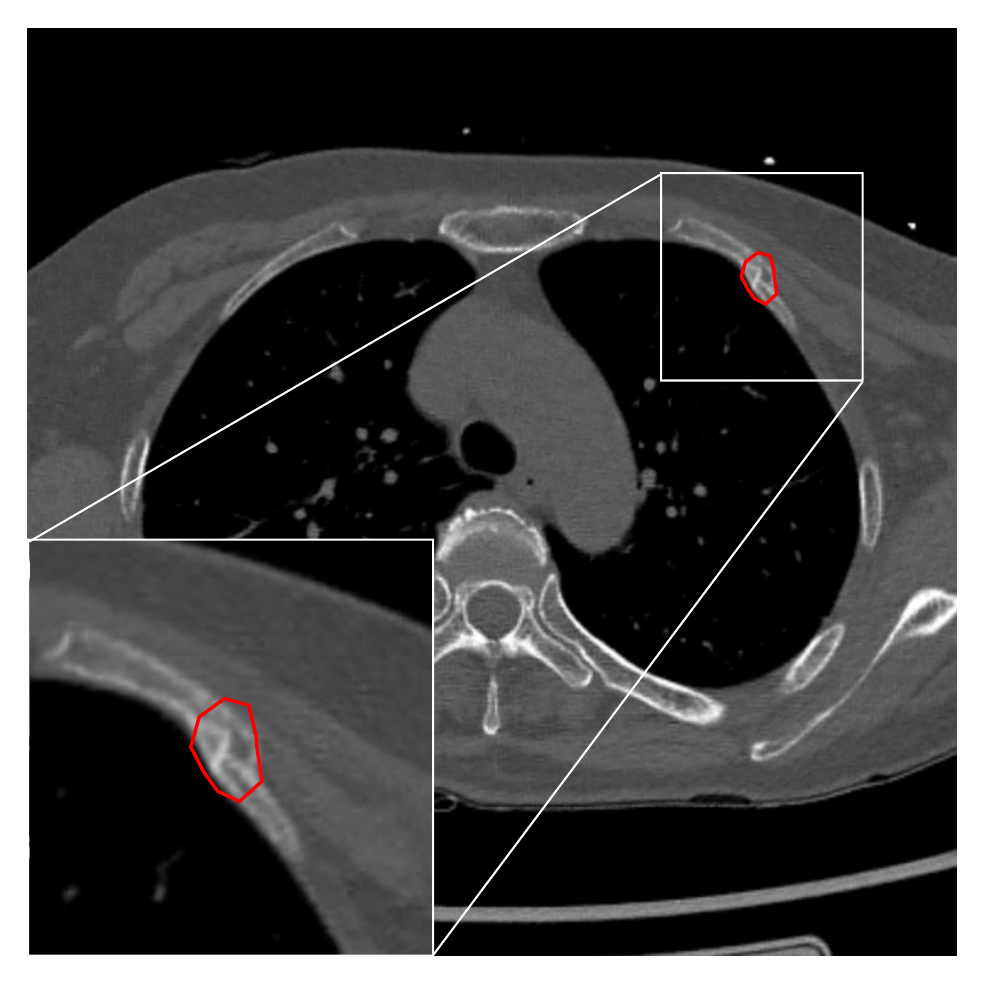}
  \caption{Mask-labeled CT}
  \label{fig:b}
\end{subfigure} \\
\begin{subfigure}{0.47\columnwidth}
  \centering
  \includegraphics[width=\linewidth]{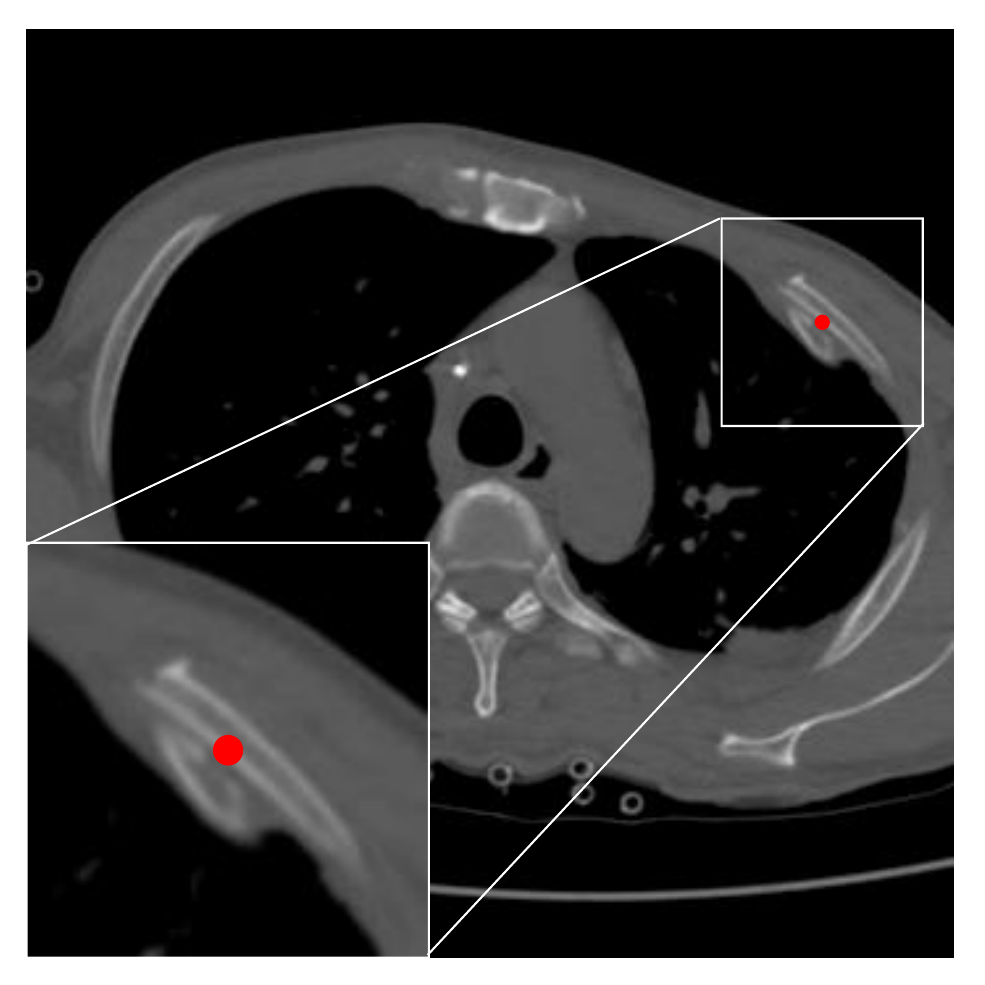}
  \caption{Dot-labeled CT}
  \label{fig:c}
\end{subfigure}
\begin{subfigure}{0.47\columnwidth}
  \centering
  \includegraphics[width=\linewidth]{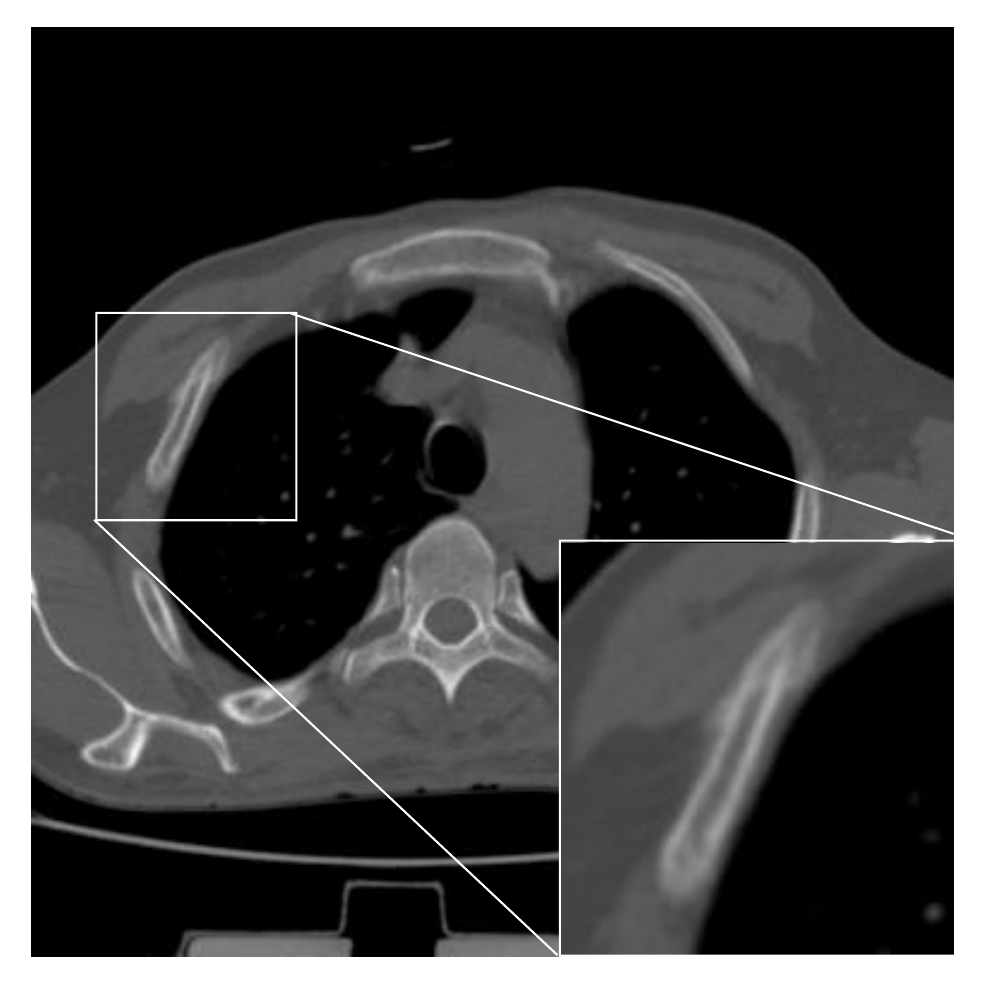}
  \caption{Unlabeled CT}
  \label{fig:d}
\end{subfigure}
\caption{Medical data can have fine-grained annotations, such as (a) boxes and (b) masks, coarse-grained annotations (or weak annotations), such as (c) dots, and the data are more often (d) unlabeled. Rib fractures are highlighted by zooming in.}
\label{fig: annotation type}
\end{figure}

Recently, deep learning (DL) has shown comparable performance to experienced radiologists on rib fracture detection \cite{weikert2020assessment,wu2021development,zhou2020automatic,luo2022rethinking}.
However, these studies rely on a large number of fine-grained annotations (e.g., lesion bounding boxes or masks) on rib fractures, which is labor-intensive and expertise-demanding.
To alleviate the labeling burden, weakly-supervised or semi-supervised algorithms have been proposed to leverage data that can be acquired more easily to improve the detection performance under a limited annotation budget \cite{jeong2019consistency,liu2021unbiased,sohn2020simple}.
Typically, weakly-supervised object detection utilizes coarser-grained labels, such as image-level labels or dot labels, which allows more efficient labeling policies \cite{zhang2021weakly}.
Semi-supervised object detection combines data with fine-grained labels and unlabeled data, which can improve detection accuracy without further labeling efforts \cite{jeong2019consistency,sohn2020simple,liu2021unbiased}.

Despite the previous efforts to develop algorithms with less fine-grained labels, practical applications are usually faced with various forms of annotations, especially for medical data.
Taking rib fractures on a computed tomography (CT) scan as an example, Fig. \ref{fig: annotation type} shows that the lesion can be box-labeled, mask-labeled, dot-labeled, or unlabeled, given varied labeling criteria and budgets across different clinical centers.
To take advantage of as much available supervision as possible, omni-supervised learning was proposed to develop unified frameworks that can be learned from data with annotations of various granularities. 
In general, existing omni-supervised object detection methods \cite{ren2020ufo,luo2021oxnet,wang2022omni} were built on generating pseudo labels.
For instance, Luo et al. \cite{luo2021oxnet} proposed a student-teacher framework that utilized a well-initialized teacher model to generate pseudo bounding boxes from weakly-labeled or unlabeled data to guide the learning of a student model.

However, the previous methods could introduce unnecessary false label assignment as the lesions do not have clear boundaries. 
From the perspective of a dense prediction task, each pixel is regarded as a training sample, and object detection often requires carefully designed label assignment for each training sample.
Hence, there is no guarantee that all samples can be clearly divided into positives or negatives given even the mask labels (which are actually polygons in practice) for the rib fractures.
As a result, the pseudo bounding box-based methods cannot provide precise and robust supervision signals to guide the learning on weakly-labeled or unlabeled data.
To tackle the aforementioned challenge, we propose co-training-guided label assignment strategies for omni-supervised learning, which eliminates the need to generate pseudo bounding boxes as well as enable robust learning from weakly-labeled and unlabeled data.

\textcolor{black}{In our previous work, we have introduced ORF-Net \cite{chai2022orf}, a framework that can utilize different granularities of supervision through an omni-supervised detection head.
In this work, we proposed ORF-Netv2 to further improve the omni-supervised learning framework.
Specifically, compared with ORF-Net, we (1) improved the label assignment strategy from ORF-Net to co-training-based dynamic label assignment, where the positive and negative attribute of a pixel is automatically learned by the model itself instead of being manually assigned; and (2) extended ORF-Net to be compatible with more types of annotations by adding a mask branch as well as specifically designing the losses for this branch.
We further conducted extensive experiments with two large-scale thoracic CT datasets and a chest X-ray dataset, demonstrating consistent improvement of ORF-Netv2 over other competitive label-efficient approaches including ORF-Net.
Moreover, we conducted budget-aware experiments to uncover the best labeling policy under limited annotation budgets based on the flexible architecture of ORF-Netv2. 
}
Our contributions can be summarized as follows:
\begin{itemize}
    \item We proposed ORF-Netv2, a novel omni-supervised rib fracture detection network supporting simultaneously learning from data with various annotation granularities.
    \item We introduced a group of novel co-training-guided label assignment strategies, which provided flexible and robust learning of the fully-labeled, weakly-labeled, and unlabeled data.
    \item Extensive experiments and analyses on three rib fracture datasets from chest radiology images demonstrated the effectiveness of our method in exploiting various granularities of annotations.
\end{itemize}

The remainder of this paper is organized as follows. We review the related literature in Section~\ref{related_work} and elaborate on the proposed method in Section~\ref{method}. We present experimental results in Section~\ref{experiments} and finally conclude in Section~\ref{conclusion}.

\section{Related Works} \label{related_work}

\subsection{Label-Efficient Object Detection}
To reduce the dependency of object detection models on fine-grained annotations, label-efficient learning \cite{jin2023label} has recently received much attention.

\textbf{Weakly-supervised learning (WSL)} utilizes labels which are not exactly the task needed.
WSL-based object detection generally uses image tags or points for model development.
For instance, WSDDN proposed a two-stream network that simultaneously learned classification and localization using image tags \cite{bilen2016weakly}.
Yang et. al introduced a framework that jointly optimized a multiple instance learning detector and a box regressor in an end-to-end manner \cite{yang2019towards}. 
More recently, some studies proposed to jointly exploit fully-labeled data and weakly-labeled data to train models. 
For example, Point DETR proposed a dot encoder applied to dot annotations, which established a one-to-one correspondence between dot annotations and objects \cite{chen2021points}. 
WSSOD \cite{fang2021wssod} introduced a pipeline that exploited the fully-labeled data with bounding boxes and weakly-labeled data with multiple image-level labels.

\textbf{Semi-supervised learning (SSL)} generally utilizes fully-annotated data together with unlabeled data. 
In object detection, SSL can be roughly categorized into consistency-based and pseudo label-based methods. 
The consistency-based methods inject consistency regularization on unlabeled data, encouraging producing robust predictions for different perturbated versions of the same data. 
For example, CSD \cite{jeong2019consistency} introduced a regularization that the model should have symmetric predictions for the images and their flipped versions. Tang et al. \cite{tang2021proposal} proposed a proposal learning module with consistency regularization on both bounding box classification and regression predictions.
Meanwhile, the pseudo label-based methods often ceased a "teacher" to generate reasonable pseudo bounding boxes to guide the target model.
For example, STAC \cite{sohn2020simple} leveraged high-confidence pseudo labels of the unlabeled images to train the model. 
Unbiased Teacher \cite{liu2021unbiased} adopted focal loss \cite{lin2017focal} to address the class imbalance caused by pseudo labels in SSL-based object detection. 

\textbf{Omni-supervised learning (OSL)} aims to simultaneously utilize different granularities of supervision.
For example, {UFO$^2$} \cite{ren2020ufo} proposed a unified framework that learned different kinds of supervision in a multi-task manner, based on a careful proposal refinement on the weakly-labeled or unlabeled data to reject false-positive proposals.
OXnet \cite{luo2021oxnet} is an omni-supervised model for chest X-ray disease detection, which unified the box supervision and the image-level supervision with a dual attention mechanism and utilized a soft focal loss to learn from unlabeled data with a teacher model.
More recently, Omni-DETR \cite{wang2022omni} introduced an omni-supervised end-to-end Transformer architecture with the student-teacher framework, which supervised the model by generating pseudo labels for different weak labels through a bipartite matching-based filtering mechanism.

\textcolor{black}{Although WSL and SSL can largely reduce labeling costs, they still lack the ability to simultaneously utilize all types of supervision.
To address the challenge, OSL was proposed to unify the learning from fully-labeled data, weakly-labeled data, and unlabeled data.
Nevertheless, there is only a dearth of OSL-based methods, and they were all based on pseudo labels, which cannot provide precise supervision signals for learning to detect lesions without clear boundaries, such as rib fractures.
This paper is one of the pioneered OSL-based detection works and the first to our knowledge to seamlessly incorporate dynamic label assignment with OSL to alleviate the reliance on pseudo bounding box labels.
}

\subsection{Label Assignment in Object Detection}
Determining positive and negative pixels in an image is a fundamental step, called label assignment, for object detection.
The label assignment strategy of current object detection methods can be categorized into two groups: fixed label assignment and dynamic label assignment. 
The fixed label assignment-based methods adopt hand-crafted rules to sample the positives and negatives during the training stage.
For example, Faster-RCNN \cite{ren2015faster} assigned labels for proposals generated by the region proposal network with predefined IoU thresholds.
FCOS \cite{tian2020fcos} took the pixels close to the center of the object bounding box to be positive samples, and others to be negative samples or ignored during training.
However, the bounding boxes cannot describe clearly the object boundaries and such a hard assignment strategy could raise many false positives or negatives. 
The dynamic assignment was hence introduced to automatically define the pixel labels. 
AutoAssign \cite{zhu2020autoassign} utilized an adaptive weighting mechanism to dynamically assign weights for each anchor by estimating the consistency metrics between its classification and localization scores. 
Recently, Li et al. \cite{li2022dual} proposed a dual-weight label assignment scheme that dynamically assigned positive and negative weights to each anchor by estimating consistency and inconsistency metrics.
\textcolor{black}{Nevertheless, these methods focused on fully-supervised object detection and cannot be easily adapted to situations where fully-labeled data are rare.
In this work, we proposed a novel co-training-guided learning scheme and further extended the dynamic label assignment to omni-supervised detection.}

\subsection{Label-efficient Object Detection in Medical Images}
Compared with natural images, the acquisition cost of fine-grained data in the medical image is more expensive due to that the annotation of lesions requires professional medical knowledge and rich experience in clinical diagnosis. 
Recently, label-efficient learning is widely used in medical image analysis to address the lack of finely annotated data \cite{jin2023label,luo2022rethinking}. 
Wang et al. \cite{wang2021knowledge} proposed an adaptive asymmetric label sharping scheme to improve the effectiveness of knowledge distillation from image-level labeled data for the task of fracture detection in chest X-rays. 
Chai et al.\cite{chai2022deep} proposed a semi-supervised framework based on deep metric learning for cervical cancer cell detection.
Bakalo et al. \cite{bakalo2021weakly} presented a deep learning architecture capable of localizing and classifying medical abnormalities in mammograms under both weakly- and semi-supervised settings. 
Wang et al. \cite{wang2020focalmix} introduced a 3D semi-supervised detection framework that utilized the unlabeled data to boost the lesions detection performance in CT scans.
\textcolor{black}{Although the above methods have reduced the model's dependence on a large amount of fully-labeled data, there are few existing works on leveraging a variety of granularities of annotations.
}

\section{Methodology}\label{method}
In this section, we first introduce the overview of the omni-supervised rib fracture detection framework. Then, we introduce the co-training-guided label assignment strategies for data in different annotation forms. At last, we describe how we train the network with different supervision signals.

\begin{figure*}[!t]
\begin{center}
	\includegraphics[width=1. \linewidth]{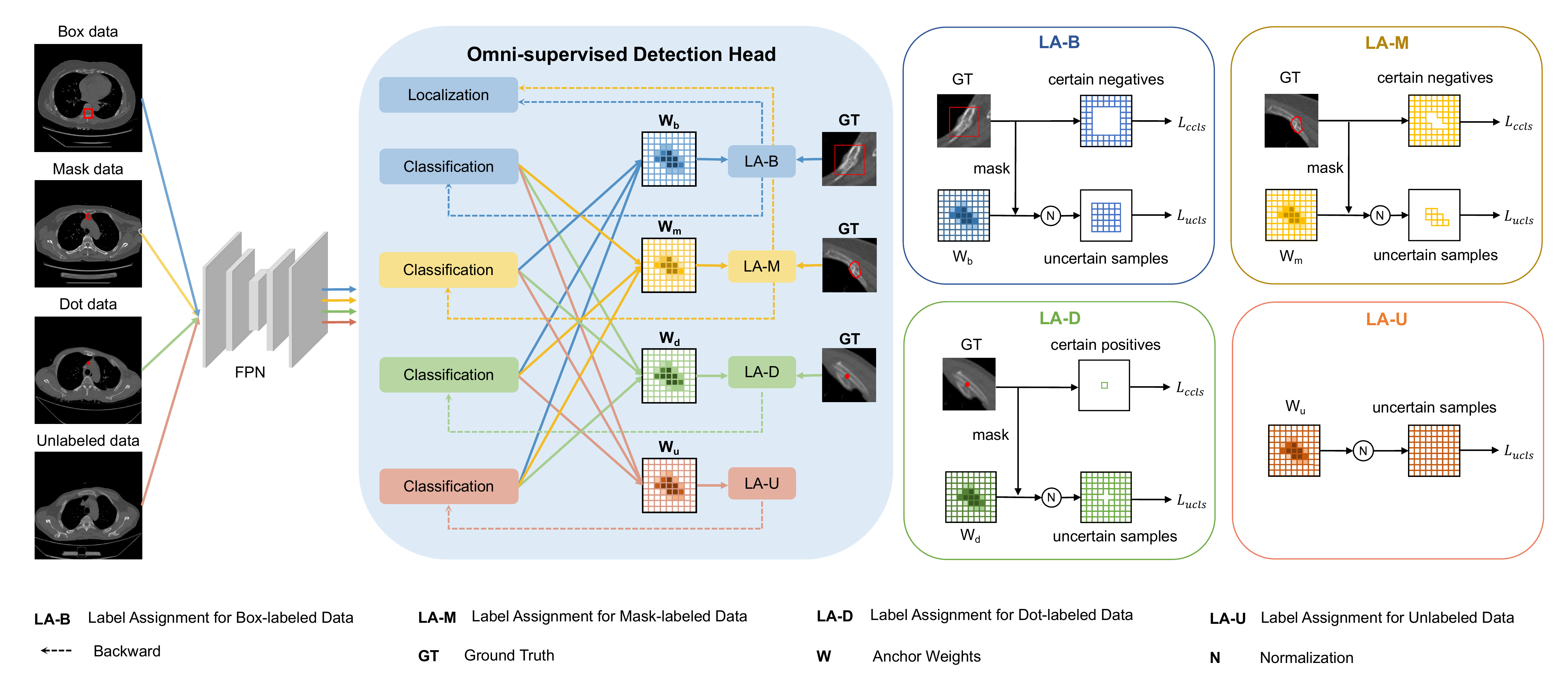}
\end{center}
\caption{Schematic view of our proposed framework. The network consists of a Feature Pyramid Network (FPN \cite{lin2017feature}) as the backbone, and an omni-supervised detection head to predict the classification score and localization information. For each form of annotated data, there is a corresponding classification branch that is trained using a dynamic label assignment strategy. }
\label{framework}
\end{figure*}

\subsection{Overview of ORF-Netv2} \label{A}
Our goal is to develop an object detector for rib fracture unifying the data with annotation of various granularities.
Considering the general annotation types for rib fracture, we have a box-labeled dataset $\mathcal{D}_{b}$ with a bounding box for each fracture, a mask-labeled dataset $\mathcal{D}_{m}$ where each fracture is with a polygon mask, a dot-labeled dataset $\mathcal{D}_{d}$ using a single dot to label each fracture, and an unlabeled dataset $\mathcal{D}_{u}$.
We propose a framework that can support training based on an arbitrary mixing of any of the above data.

The framework of the proposed omni-supervised rib fracture detector, ORF-Netv2, is illustrated in Fig. \ref{framework}.
ORF-Netv2 is based on FCOS \cite{tian2020fcos}, an anchor-free object detector that learns in a fully-convolutional per-pixel prediction manner.
Specifically, the proposed network first extracts the rich multi-layer pyramid features $X^{\rm fpn}$ using the Feature Pyramid Network (FPN \cite{lin2017feature}), and then performs object classification and localization with a novel omni-supervised detection head.
The omni-supervised detection head contains a localization branch for bounding box regressing and multiple parallel classification branches.
Particularly, each classification branch is supervised by a certain type of data.
Following \cite{tian2020fcos}, the classification branches and regression branch consist of five convolutional layers, including the last prediction layer.

\subsection{Co-training-based Sample Weighting} \label{B}
We notice that the pixel samples for all types of data can be divided into two classes: certain samples and uncertain samples. 
The certain samples can be clearly defined as positive or negative by the annotations.
For example, the pixels outside the bounding boxes and masks are certain negatives, and the pixels labeled with dots are certain positives.
The uncertain samples cannot be clearly defined due to the ambiguity of annotations.
Specifically, these samples can be those inside the bounding boxes or the coarse masks, those beyond the labeled dots, or those on the unlabeled data.
As a result, the general fixed label assignment strategy could cause many false positives and negatives, and cannot be flexibly transferred to learn the weakly-labeled or unlabeled data.

Dynamic label assignment can alleviate the challenge of ambiguous annotations by learning an automatic sample weighting policy.
However, many existing methods \cite{liu2020hambox, ke2020multiple, kim2020probabilistic, feng2021tood} were based on using self-predicted confidence scores as the indicators for label assignment, which could lead to overfitting \cite{chai2022orf}.
To tackle this challenge, we take advantage of the multi-branch structure and propose the co-training-based label assignment strategy.
Specifically, co-training \cite{blum1998combining} minimizes the divergence of two learning algorithms trained on two different views of the same data.
As the branches of our omni-supervised detection head are trained with different data, the predictions by these branches would also be divergent.
In light of this observation, we generate the inter-guided map $I$ from other branches to guide the learning of the current branch.

Formally, we define the outputs of the box-labeled branch, the mask-labeled branch, the dot-labeled branch, and the unlabeled branch as ${P}_{b}$, ${P}_{m}$, ${P}_{d}$, ${P}_{u}$, respectively. 
The inter-guided map $I$ for each branch is calculated as follows:
\begin{equation}
\begin{split}
    {I}_{b} = ({P}_{m}\times{P}_{d}\times{P}_{u})^\frac{1}{3}, {I}_{m} = ({P}_{b}\times{P}_{d}\times{P}_{u})^\frac{1}{3}, \\
    {I}_{d} = ({P}_{b}\times{P}_{m}\times{P}_{u})^\frac{1}{3} , {I}_{u} = ({P}_{b}\times{P}_{m}\times{P}_{d})^\frac{1}{3}. \label{ia:inter-label assignment}
\end{split}
\end{equation}

The inter-guided maps reveal the agreement on the probabilities of different branches, which can be used as a more reliable indicator on which the values represent the confidences of label assignment.
We then use the maps to assign weights for the pixels to indicate their importance during the learning process.
Specifically, for dot-labeled data and unlabeled data, the inter-guided map $I$ is normalized as follows:
\begin{equation}
\begin{split}
    {W}_{d} = N({I}_{d}), {W}_{u} = N({I}_{u}). \label{ia:dot-unlabel metric}
\end{split}
\end{equation}
\noindent where $N(\cdot)$ represents a linear normalization function to scale the maps into $[0, 1]$.

Further, a positive training pixel sample not only obtains a high classification score but also locates in an accurate position.
Therefore, for the box-labeled and mask-labeled data of which the annotations contain more precise location information, we also take the ground truth in our dynamic label assignment. 
Specifically, we combine the inter-guided maps $I$ and the intersection over union (IoU) scores between the predicted boxes and the ground truth to generate more reliable weights.
The sample weights for pixels on the box-labeled and mask-labeled data can be obtained as follows:
\begin{equation}
\begin{split}
    {W}_{b} = N(({I}_{b})^{\alpha}\times({IoU}_{b})^{\beta}), \\
    {W}_{m} = N(({I}_{m})^{\alpha}\times({IoU}_{m})^{\beta}), \label{ia:box-mask metric}
\end{split}
\end{equation}
where $\alpha$ and $\beta$ are used to balance the contributions of classification confidence and the IoU score.
For mask annotations, the IoU scores are computed using their bounding boxes.

\subsection{Co-training-guided Dynamic Label Assignment} \label{C}

The increase or decrease of sample weights essentially reveals the model's confidence of a sample being positive or negative, respectively.
Hence, for each uncertain sample, we can apply the co-training-based weights to the objectives of classification branches to dynamically assign their labels.
Specifically, we propose three label assignment strategies: hard label assignment, soft label assignment, and dynamic label assignment.

\noindent\textbf{Hard Label Assignment:} A predefined threshold $t$ is used to divide the pixels into positive or negative samples.
Specifically, denote $P$ as the output probability of one of the classification branches in ORF-Netv2, $W$ as the corresponding sample weights obtained before, $i$ as the index for the pixel on the FPN feature map, and $S$ as the total number of pixel samples, the training objective is computed as follows:

\begin{equation}
\mathcal{L}_{cls}= 
\left\{ 
    \begin{array}{lc}
        -\sum\limits_{i}\limits^S(1-P^i)^{\gamma}\log{P^i}, & W^i\geq t \\[8pt]
        -\sum\limits_{i}\limits^S(P^i)^{\gamma}\log{(1-P^i)}, & W^i<t \\
    \end{array}
\right.
\end{equation}
\noindent where the Focal loss \cite{lin2017focal} is adopted based on the positive or negative samples assigned by $W$.

\noindent\textbf{Soft Label Assignment:}
The hard label assignment helps distinguish the positive samples from the negative ones. 
Nevertheless, with a fixed threshold defining the samples, there still could be false label assignments.
As mentioned, the sample weights can be also regarded as indicators to measure the importance of training samples. 
Therefore, we propose the soft label assignment strategy which uses $W$ to further emphasize the samples with a high confidence score during training and pay less attention to the low-confidence samples. 
The weighted loss functions for positive samples and negative samples are derived as follows:

\begin{equation}
\mathcal{L}_{cls}= 
\left\{ 
    \begin{array}{lc}
        -\sum\limits_{i}\limits^S(W^i)^{\gamma}(1-P^i)^{\gamma}\log{\left((1-W^i)P^i\right)},W^i\geq t \\[8pt]
        -\sum\limits_{i}\limits^S(1-W^i)^{\gamma}(P^i)^{\gamma}\log{\left(W^i(1-P^i)\right)},W^i<t \\
    \end{array}
\right.
\end{equation}

Here, $W$ is used to weigh the focal loss, so that the model could determine the importance of a training sample and alleviate the potential false label assignment.

\noindent\textbf{Dynamical Label Assignment:}
We notice that $W$ could vary among different branches and change with the training of the model.
As a result, there is no guarantee that the threshold $t$ can be a general choice.
To tackle this challenge, instead of defining positive or negative samples, we allow the model to learn to dynamically adjust the objective of each sample.
We propose the dynamic label assignment with the following:

\begin{equation}
\begin{split}
    \mathcal{L}_{cls} = -\sum_{i}^S(W^{i})^{\gamma}(1-P^{i})^{\gamma}\log{\left((1-W^i)P^{i}\right)}\\
    +(1-W^{i})^{\gamma}(P^{i})^{\gamma}\log{\left(W^i(1-P^{i})\right)}.
\end{split}
\label{eq:dla}
\end{equation}
\noindent where the model is set to learn a unified objective for an uncertain sample. 
With the increase or decrease of $W$, the above loss could dynamically determine a pixel to be more likely a positive sample or a negative sample, respectively.

\subsection{Omni-Supervision for Different Annotation Data} \label{D}

We train ORF-Netv2 based on the proposed co-training-guided label assignment strategies.
Using dynamic label assignment as an example, we here show how we enable learning under different supervision.

For the certain samples, i.e., the negative samples outside bounding-boxes and masks as well as the positive samples labeled with dots, we adopt the Focal loss as follows:

\begin{equation}
\mathcal{L}_{ccls}= 
\left\{ 
    \begin{array}{lc}
        -\sum\limits_{i}\limits^{S}(1-P^i)^{\gamma}\log{P^i}, & i\in positives \\[8pt]
        -\sum\limits_{i}\limits^{S}(P^i)^{\gamma}\log{(1-P^i)}, & i\in negatives \\
    \end{array}
\right.
\end{equation}

For the uncertain samples, i.e., the samples inside the bounding boxes or the coarse masks, the samples outside the dot labels, and the samples from the unlabeled data, we utilize our proposed dynamic label assignment from Eq. \ref{eq:dla} and set the objective as follows:
\begin{equation}
\begin{split}
    \mathcal{L}_{ucls} = -\sum_{j}^{N}\sum_{i}^{M^{j}} (W^{ij})^{\gamma}(1-P^{ij})^{\gamma}\log{\left((1-W^{ij})P^{ij}\right)}\\
    +(1-W^{ij})^{\gamma}(P^{ij})^{\gamma}\log{\left(W^{ij}(1-P^{ij})\right)},
\end{split}
\end{equation}
where $W$ denotes the co-training-guided sample weights, $N$ denotes the number of uncertain regions, and $M^{j}$ is the number of samples in the $j$-th region. For the box-labeled data or mask-labeled data, $N$ is the number of boxes or masks. For the dot-labeled data or unlabeled data, $N=1$.

Moreover, we use the generalized IoU (GIoU) loss \cite{rezatofighi2019generalized} to train the localization branch based on the box-labeled and mask-labeled data:
\begin{equation}
    \mathcal{L}_{reg} = \sum_{j}^{N}\sum_{i}^{M^{j}}{\mathcal{L}_{\rm GIoU}(h^i,\hat{h}^i)},
\end{equation} 
where ${h}$ denotes the predicted bounding boxes and $\hat{h}$ denotes the corresponding ground-truth boxes.

The overall loss function for ORF-Netv2 is as follows:
\begin{equation}
\begin{split}
    \mathcal{L} = (\mathcal{L}_{ucls}^b + \mathcal{L}_{ccls}^b+\mathcal{L}_{reg}^{b})+(\mathcal{L}_{ucls}^m + \mathcal{L}_{ccls}^m+\mathcal{L}_{reg}^{m}) + \\
    (\mathcal{L}_{ucls}^d+ \mathcal{L}_{ccls}^d) +\delta(\mathcal{L}_{ucls}^u)
\end{split}
\end{equation}
where $\delta$ is a hyper-parameter to weigh and stabilize the training of the unsupervised classification branch. 
$\mathcal{L}^b$, $\mathcal{L}^m$, $\mathcal{L}^d$, and $\mathcal{L}^u$ represent the loss for the box-supervised, mask-supervised, dot-supervised, or unsupervised branch, respectively.
During training, each classification branch receives the supervision signal from a specific type of data, e.g., $\mathcal{L}^d$ will only be computed based on the dot-labeled data, and the remaining classification branches will assist in determining the sample weights with inter-guided maps.
Also, the gradients will not be propagated back through the inter-guided map, and hence the other branches won't be trained.
The localization branch is trained with the box-labeled data and the mask-labeled data.
During testing, we simply take the average results of the classification branches and combine them with the result from the localization branch to generate the final detection results.

\section{Experiments}\label{experiments}
\subsection{Datasets}
\textbf{RibFrac:} The RibFrac \cite{jin2020deep} dataset contains 500 chest-abdomen CT scans from patients with traumatic rib fractures. These scans were first diagnosed by two radiologists (3-5 years and 10-20 years of experience). Then, two radiologists (each with 5 years of experience) delineated a polygon mask for each traumatic rib fracture based on the diagnostic report, which was further confirmed by a senior radiologist (20 years of experience).
All the mask labels are at the instance level.
To study omni-supervised learning, we randomly selected 185 cases to be box-labeled, from which 105 cases (11,381 positive slices, 26,288 negative slices) were used for training, and 80 cases (5,526 positive slices, 20,814 negative slices) were used for testing.
The remaining 315 cases are used for training as well, from which 105 cases (10,886 positive slices, 27,674 negative slices) were mask-labeled, 105 cases (11,143 positive slices, 27,745 negative slices) were dot-labeled, and 105 cases (38,353 slices) were unlabeled.
The bounding boxes of the original mask annotations were generated to be box annotations, and the center dots of the masks were used as the dot annotations.

\textbf{CRF:} The CRF dataset \cite{chai2022orf} is an in-house dataset with 2,239 chest CT scans collected from multiple hospitals.
This dataset naturally contains bounding box labels, dot labels, as well as unlabeled data.
Specifically, there were in total 685 cases labeled in boxes, from which 224 (8,264 positive slices, 57,490 negative slices) were used for training, 151 (4,999 positive slices, 43,078 negative slices) were used for validation, and 310 (12,689 positive slices, 91,227 negative slices) were used for testing.
Meanwhile, there were 450 scans labeled in dot (22,328 positive slices, 186,485 negative slices) and 1,104 scans unlabeled (338,644 slices), which are all used for training.
The boxes and dots were first provided by a radiologist (10 years of experience) and then checked by a senior radiologist (18 years of experience). 
\textcolor{black}{CRF does not contain mask annotations.}

\textbf{XRF:} The XRF dataset is an in-house dataset that includes a total of 8,328 chest X-rays (CXRs). A total of 10 radiologists (4-30 years of experience) were involved in marking the bounding boxes of the rib fractures. Each image has a corresponding text report and is labeled by two physicians. If the initial annotators disagreed with each other, a final decision was made by a senior radiologist ($\ge$ 20 years of experience).
We randomly split the data into a training set and a testing set.
The training set consisted of 6,362 CXRs, from which 1,185 contained fractures (395 labeled w/ boxes, 395 labeled w/ dots, 395 unlabeled), and the remaining were normal.
The testing set contained 1,966 CXRs, from which 153 were positive cases labeled with boxes and 1813 were negative cases.
\textcolor{black}{XRF does not contain mask annotations.}

We took 2D slices as the inputs by scaling each CT slice to 1024*1024 and then constructing three-channel images by copying the original slice. Finally, the input images are of dimension 1024$\times$1024$\times$3.

\subsection{Implementation Details and Evaluation Metrics}
We used FCOS with ResNet-50 \cite{he2016deep} backbone pre-trained from ImageNet \cite{deng2009imagenet} as our base model.
All models involved in our experiments were implemented based on Pytorch \cite{paszke2019pytorch} \textcolor{black}{and a TITAN Xp GPU.} 
Note that we also used a base model trained with only the box-labeled data to select slices with potential fractures from the unlabeled dataset for later model development.
During training, we equally sampled the different types of data.
Horizontal flipping was performed to augment the training data.
\textcolor{black}{For all experiments, we trained the models for 70000 iterations for better convergence.
During the training stage, we used the validation sets to test the model every 2500 iterations, and the model that performed best on the validation set will be selected as the final model.
}
Stochastic Gradient Descent (SGD) with a momentum of 0.9 was employed.
The initial learning rate was set to 0.001 and then divided by 10 every 30000 iterations.
The threshold $t$ in the hard label assignment strategy and the soft label assignment strategy was empirically set as 0.5.
We set $\delta$ as the max value of $I_u$ to weigh the unlabeled loss.

During testing, we used the bounding boxes of fractures for evaluation.
The non-maximum suppression (NMS) with an IoU threshold of 0.6 was used for post-processing in all experiments.
\textcolor{black}{We used the common evaluation COCO API for model evaluation \footnote{\url{https://cocodataset.org/\#detection-eval}}.
}
Considering the small sizes of fractures, we used the mean Average Precision (mAP) from AP40 to AP75 with an interval of 5 and AP50 as the evaluation metrics.
\textcolor{black}{As most of the rib fracture lesions are small targets with bounding boxes smaller than 32$\times$32, we did not compute the APs, APm, and APl for further evaluation.
To enhance the comparison with the state-of-the-arts, we used bootstrapping (1000 times of sampling) for the results by each method and computed the p-values between the bootstrapped results with paired t-test.
}

\subsection{Ablation Study}

\subsubsection{Effectiveness of label assignment strategies}
As mentioned, label assignment strategy plays an important role in object detection.
Here, we compared the effectiveness of the Hard Label Assignment (HLA) strategy, the Soft Label Assignment (SLA) strategy, and the Dynamic Label Assignment (DLA) strategy with experiments on the RibFrac dataset.
As shown in Table \ref{ablation studies}, when all data were used, the HLA strategy achieved 33.7\% mAP and 46.8\% AP50 on the testing set. 
Meanwhile, the SLA strategy achieved 34.1\% mAP and 47.2\% AP50, showing the effectiveness of adding soft weights to the training objective. 
Moreover, our proposed DLA strategy achieved the best performance (34.7\% mAP and 48.1\% AP), clearly surpassing the other label assignment strategies with 1.\% mAP and 1.2\% AP50 higher than HLA and 0.6\% mAP and 0.9\% AP higher than SLA.
These results demonstrate the improvement brought by the great flexibility offered by dynamic label assignment.

\begin{table}[t] \caption {Comparison of different label assignment strategies on RibFrac.}
\centering
\begin{tabular}{ccccccc}
\hline
\multirow{2}{*}{Method} & \multicolumn{4}{c}{\#scans used}                                                                        &\multicolumn{2}{c}{Metrics} \\ \cline{2-7} 
                        & \multicolumn{1}{c}{\begin{tabular}[c]{@{}c@{}}$D_b$\end{tabular}}
                        & \multicolumn{1}{c}{\begin{tabular}[c]{@{}c@{}}$D_m$\end{tabular}}& \multicolumn{1}{c}{\begin{tabular}[c]{@{}c@{}}$D_d$\end{tabular}} & \begin{tabular}[c]{@{}c@{}}$D_u$\end{tabular} & mAP          & AP50 \\ \hline

HLA                & \multicolumn{1}{c}{105}                                                         & \multicolumn{1}{c}{105}             &\multicolumn{1}{c}{105}                                             & 105                                                      &33.7         & 46.8         \\ \hline

SLA                & \multicolumn{1}{c}{105}                                                         & \multicolumn{1}{c}{105}             &\multicolumn{1}{c}{105}                                             & 105                                                      &34.1         & 47.2        \\ \hline

DLA                & \multicolumn{1}{c}{105}                                                         & \multicolumn{1}{c}{105}             &\multicolumn{1}{c}{105}                                             & 105                                                      &\textbf{34.7}         & \textbf{48.1}         \\ \hline

\end{tabular}
\label{ablation studies}
\end{table}

\subsubsection{Analysis of classification branches}
\begin{figure}[ht]
\begin{center}
	\includegraphics[width=1. \linewidth]{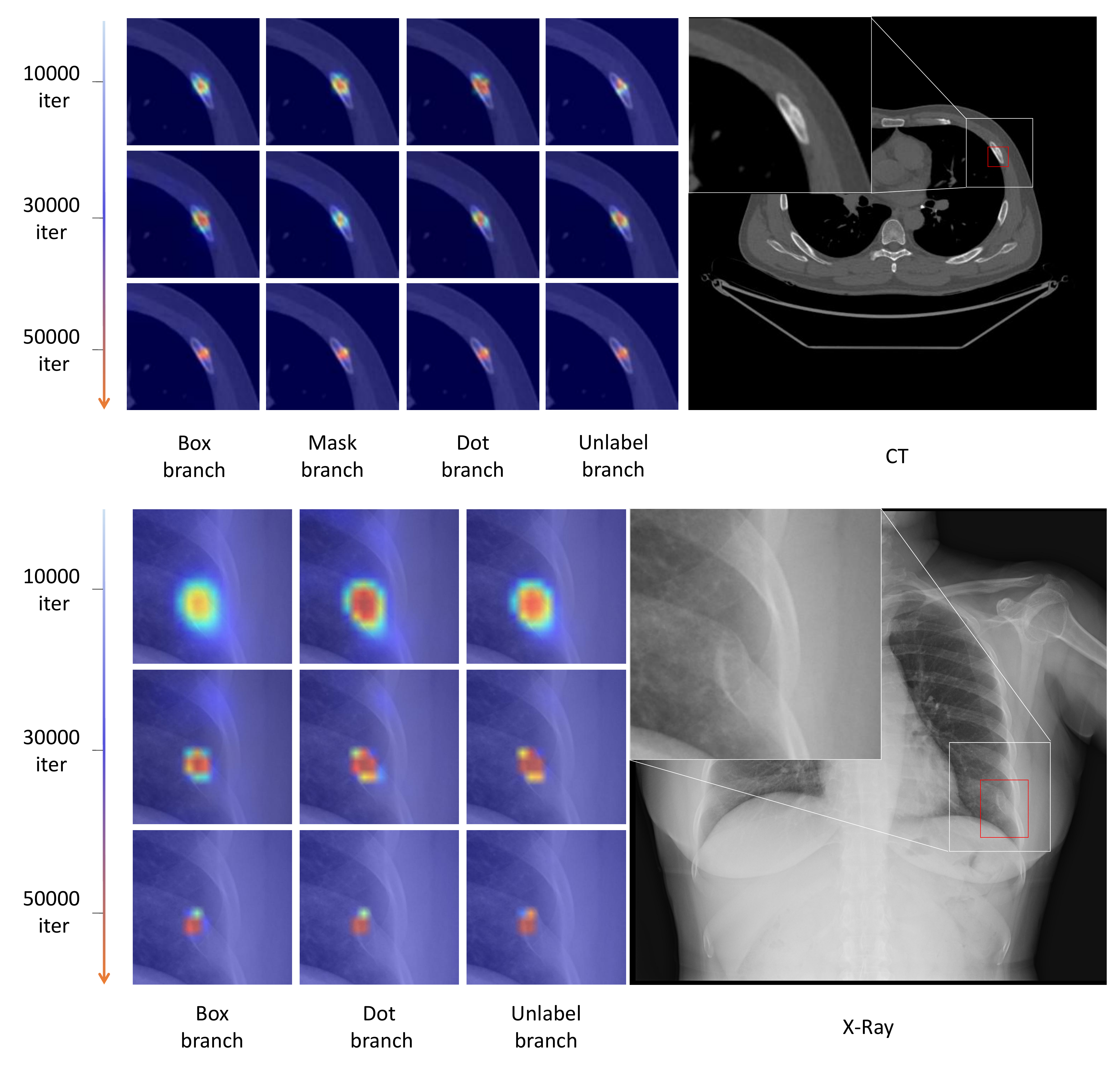}
\end{center}
\caption{Visualization of the predicted maps from different classification branches at different iteration numbers.}
\label{fig:LA}
\end{figure}

We report in Table \ref{tab: cls branches} the performance of the different classification branches.
Note that the same localization branch was used to generate the detection results.
It can be found that the mAP performance of different classification branches fluctuates slightly between 34.3\% and 34.7\%, which demonstrates that the proposed co-training-based label assignment strategy could prompt the different branches to maximize their agreement rib fracture detection. 
\textcolor{black}{We also illustrate in Fig. \ref{fig:LA} the predicted maps from each classification branch.
The visualization shows that as the training iteration increases, the prediction maps of different classification branches become more accurate and more consistent with each other.
Both the quantitative and qualitative results demonstrate that our proposed co-training-based dynamic label assignment strategy can effectively foster mutual learning between the branches, despite that they were trained with different data.
Co-training finally led to comparable results for each branch, and we fuse the results from different branches as a more robust and accurate strategy. 
}

\begin{table}[t] \caption {Performance comparison of different classification branches in the omni-supervised detection head on RibFrac.}
\centering
\begin{tabular}{ccccccc}
\hline
\multirow{2}{*}{Method} & \multicolumn{4}{c}{\#scans used}                                                                        &\multicolumn{2}{c}{Metrics} \\ \cline{2-7} 
                        & \multicolumn{1}{c}{\begin{tabular}[c]{@{}c@{}}$D_b$\end{tabular}}
                        & \multicolumn{1}{c}{\begin{tabular}[c]{@{}c@{}}$D_m$\end{tabular}}& \multicolumn{1}{c}{\begin{tabular}[c]{@{}c@{}}$D_d$\end{tabular}} & \begin{tabular}[c]{@{}c@{}}$D_u$\end{tabular} & mAP          & AP50           \\ \hline

Box branch                & \multicolumn{1}{c}{105}                                                         & \multicolumn{1}{c}{105}             &\multicolumn{1}{c}{105}                                             & 105                                                      &34.6         & 48.0         \\ \hline

Mask branch                & \multicolumn{1}{c}{105}                                                         & \multicolumn{1}{c}{105}             &\multicolumn{1}{c}{105}                                             & 105                                                      &34.4         & 47.7        \\ \hline

Dot branch                & \multicolumn{1}{c}{105}                                                         & \multicolumn{1}{c}{105}             &\multicolumn{1}{c}{105}                                             & 105                                                      &34.3         & 47.6         \\ \hline

Unlabeled branch                & \multicolumn{1}{c}{105}                                                         & \multicolumn{1}{c}{105}             &\multicolumn{1}{c}{105}                                             & 105                                                      &34.6         & \textbf{48.1}        \\ \hline
Fusion                & \multicolumn{1}{c}{105}                                                         & \multicolumn{1}{c}{105}             &\multicolumn{1}{c}{105}                                             & 105                                                      &\textbf{34.7}         & \textbf{48.1}         \\ \hline

\end{tabular}
\label{tab: cls branches}
\end{table}

\subsubsection{Impact of hyper-parameters in sample weighting}
As the localization accuracy is also an important factor in label assignment \cite{feng2021tood}, we combined the scores on the inter-guided map $I$ as well as the IoU scores in sample weighting for box-labeled and mask-labeled data, as in Eq. \ref{ia:box-mask metric}.
To balance the contributions to the final weights between $I$ and IoU, we introduced two hyper-parameters $\alpha$ and $\beta$.
Here, we study the impact of the two weights with results reported in Table \ref{table hyper parameters}.
With a coarse search of hyper-parameters, we observed that the best result of 34.7\% mAP and 48.1\% AP could be achieved when $\alpha$ is set the maximum score in the corresponding inter-guided map and $\beta$ set to 1. 
Other combinations of $\alpha$ and $\beta$ would degrade the mAP performance from 0.2\% to 1.9\%.
We thus adopted the best combination throughout our experiments.

\begin{table}[t] \caption {Comparison of different hyper-parameters in the sample weighting function on RibFrac.}
\centering
\begin{tabular}{ccccc}
\hline
\multirow{2}{*}{Method} & \multirow{2}{*}{$\alpha$}&  \multirow{2}{*}{$\beta$} &\multicolumn{2}{c}{Metrics} \\ \cline{4-5} &  &  & mAP          & AP50           \\ \hline
ORF-Netv2                & \multicolumn{1}{c}{1}                                                         & \multicolumn{1}{c}{1}                                                             &32.8         &46.0         \\ \hline
ORF-Netv2                & \multicolumn{1}{c}{0.5}                                                         & \multicolumn{1}{c}{1}                                                       &34.5        & 47.6 \\ \hline    
ORF-Netv2                & \multicolumn{1}{c}{MI}                                                         & \multicolumn{1}{c}{1}                                                             &\textbf{34.7}         & \textbf{48.1}        \\ \hline
ORF-Netv2                & \multicolumn{1}{c}{1}                                                         & \multicolumn{1}{c}{2}                                                             &33.6        & 46.5 \\ \hline
ORF-Netv2                & \multicolumn{1}{c}{0.5}                                                         & \multicolumn{1}{c}{2}                                                             &34.5        & 47.8 \\ \hline

\end{tabular}

\begin{tablenotes}
    \scriptsize
    \item[a] $\alpha$, $\beta$ indicate the two hyper-parameters used in the sample weights for box-labeled data and mask-labeled data. $MI$ denote the max score of the inter-guided map.
\end{tablenotes}

\label{table hyper parameters}
\end{table}

\begin{table}[t] \caption {Comparison with the SOTA on RibFrac. *: p-value $<$ 0.05; **: p-value $<$ 0.01.}
\setlength{\tabcolsep}{5pt}
\color{black}
\centering
\begin{tabular}{ccccccccc}
\hline
\multirow{2}{*}{Index} & \multirow{2}{*}{Method} & \multicolumn{4}{c}{\#scans used}                                                                        &\multicolumn{2}{c}{Metrics} & \multirow{2}{*}{P} \\ \cline{3-8} 
                     &    & \multicolumn{1}{c}{\begin{tabular}[c]{@{}c@{}}$D_b$\end{tabular}}
                        & \multicolumn{1}{c}{\begin{tabular}[c]{@{}c@{}}$D_m$\end{tabular}}& \multicolumn{1}{c}{\begin{tabular}[c]{@{}c@{}}$D_d$\end{tabular}} & \begin{tabular}[c]{@{}c@{}}$D_u$\end{tabular} & mAP           & AP50          \\ \hline
1 & FCOS \cite{tian2020fcos}                   & \multicolumn{1}{c}{105}                                                         & \multicolumn{1}{c}{0}                                  & \multicolumn{1}{c}{0}                         & 0                                                         &30.9         &42.5         &- \\ \hline

2 & FCOS \cite{tian2020fcos}        & \multicolumn{1}{c}{105}                                                         & \multicolumn{1}{c}{105}               & \multicolumn{1}{c}{0}                                          & 0                                                            & \textbf{33.8}         &46.9        &* \\

3 & ORF-Netv2                    & \multicolumn{1}{c}{105}                                                         & \multicolumn{1}{c}{105}               & \multicolumn{1}{c}{0}                                          & 0                                                      &\textbf{33.8}        & \textbf{47.3}       &-  \\ 
\hline

4 & FCOS \cite{tian2020fcos}        & \multicolumn{1}{c}{105}                                                         & \multicolumn{1}{c}{0}               & \multicolumn{1}{c}{105}                                          & 0                                                             &31.4         &43.1        &** \\
5 & ORF-Net \cite{chai2022orf}                & \multicolumn{1}{c}{105}                                                         & \multicolumn{1}{c}{0}                        & \multicolumn{1}{c}{105}                                 & 0                                                         &31.8         &44.7      &*   \\
6 & ORF-Netv2                 & \multicolumn{1}{c}{105}                                                         & \multicolumn{1}{c}{0}                        & \multicolumn{1}{c}{105}                                & 0                                                         & \textbf{32.0}         & \textbf{45.4}      &-   \\ \hline
7 & $\Pi$ Model \cite{laine2016temporal}                    & \multicolumn{1}{c}{105}                                                         & \multicolumn{1}{c}{0}                & \multicolumn{1}{c}{0}                                         & 105                                                        & 31.4        &42.5  &* \\
8 & STAC \cite{sohn2020simple}                    & \multicolumn{1}{c}{105}                                                         & \multicolumn{1}{c}{0}                & \multicolumn{1}{c}{0}                                         & 105                                                        & 31.3        &43.2  &* \\
9 & AALS  \cite{wang2021knowledge}                    & \multicolumn{1}{c}{105}                                                         & \multicolumn{1}{c}{0}                & \multicolumn{1}{c}{0}                                         & 105                                                        & \textbf{31.5}        &42.9 &** \\
10 & OXNet  \cite{luo2021oxnet}                    & \multicolumn{1}{c}{105}                                                         & \multicolumn{1}{c}{0}                & \multicolumn{1}{c}{0}                                         & 105                                                        & 31.1       &43.1 &* \\
11 &UT \cite{liu2021unbiased}                    & \multicolumn{1}{c}{105}                                                         & \multicolumn{1}{c}{0}                & \multicolumn{1}{c}{0}                                         & 105                                                        &31.4         & \textbf{43.5}  &*\\

12 &ORF-Net \cite{chai2022orf}                    & \multicolumn{1}{c}{105}                                                         & \multicolumn{1}{c}{0}                & \multicolumn{1}{c}{0}                                         & 105                                                        &31.1         &43.2 &* \\
13 & ORF-Netv2                     & \multicolumn{1}{c}{105}                                                         & \multicolumn{1}{c}{0}                & \multicolumn{1}{c}{0}                                         & 105                                                        &31.2         &43.4       &- \\
\hline
14 & $\Pi$ Model \cite{laine2016temporal}    & \multicolumn{1}{c}{105}                                                         & \multicolumn{1}{c}{0}               & \multicolumn{1}{c}{105}                                          & 105                                                      &32.8         &45.0     &**    \\
15 & STAC \cite{sohn2020simple}      & \multicolumn{1}{c}{105}                                                         & \multicolumn{1}{c}{0}               & \multicolumn{1}{c}{105}                                          & 105                                                      &\textbf{33.4}         &44.9      &*   \\

16 & AALS  \cite{wang2021knowledge}    & \multicolumn{1}{c}{105}                                                         & \multicolumn{1}{c}{0}               & \multicolumn{1}{c}{105}                                          & 105                                                      &33.3         &44.9      &*   \\
17 & OXNet  \cite{luo2021oxnet}                     & \multicolumn{1}{c}{105}                                                         & \multicolumn{1}{c}{0}               & \multicolumn{1}{c}{105}                                          & 105                                                      &33.1        &45.1     &*    \\
18 & UT \cite{liu2021unbiased}                   & \multicolumn{1}{c}{105}                                                         & \multicolumn{1}{c}{0}               & \multicolumn{1}{c}{105}                                          & 105                                                      &33.0        &45.0       &*  \\
19 & ORF-Net \cite{chai2022orf}   & \multicolumn{1}{c}{105}          & \multicolumn{1}{c}{0}               & \multicolumn{1}{c}{105}                                          & 105                                                      &32.9         &45.3        &** \\
20 & ORF-Netv2                    & \multicolumn{1}{c}{105}                                                         & \multicolumn{1}{c}{0}               & \multicolumn{1}{c}{105}                                          & 105                                                      &\textbf{33.4}         & \textbf{46.6}      &-   \\ \hline
21 & $\Pi$ Model \cite{laine2016temporal}                & \multicolumn{1}{c}{105}                                                         & \multicolumn{1}{c}{105}               & \multicolumn{1}{c}{0}                                          & 105                                                      &33.4         &46.9 &**\\
22  & STAC \cite{sohn2020simple}                  & \multicolumn{1}{c}{105}                                                         & \multicolumn{1}{c}{105}               & \multicolumn{1}{c}{0}                                          & 105                                                      &33.6         &45.9 &**\\

23 & AALS  \cite{wang2021knowledge}                 & \multicolumn{1}{c}{105}                                                         & \multicolumn{1}{c}{105}               & \multicolumn{1}{c}{0}                                          & 105                                                      &33.5         &46.8 &**\\
24& OXNet  \cite{luo2021oxnet}                  & \multicolumn{1}{c}{105}                                                         & \multicolumn{1}{c}{105}               & \multicolumn{1}{c}{0}                                          & 105                                                      &33.1         &46.9       &** \\
25 & UT  \cite{liu2021unbiased}                 & \multicolumn{1}{c}{105}                                                         & \multicolumn{1}{c}{105}               & \multicolumn{1}{c}{0}                                          & 105                                                      &33.3         &47.0       &** \\
26 & ORF-Net  \cite{chai2022orf}                   & \multicolumn{1}{c}{105}                                                         & \multicolumn{1}{c}{105}               & \multicolumn{1}{c}{0}                                          & 105                                                      &33.7         & 47.1     &**   \\
27 & ORF-Netv2                    & \multicolumn{1}{c}{105}                                                         & \multicolumn{1}{c}{105}               & \multicolumn{1}{c}{0}                                          & 105                                                      &\textbf{34.2}         & \textbf{47.4}      &-  \\ \hline
28 & $\Pi$ Model \cite{laine2016temporal}               & \multicolumn{1}{c}{105}                                                         & \multicolumn{1}{c}{105}               & \multicolumn{1}{c}{105}                                          & 105                                                      &34.0         &47.0      &**    \\
29 & STAC \cite{sohn2020simple}                 & \multicolumn{1}{c}{105}                                                         & \multicolumn{1}{c}{105}               & \multicolumn{1}{c}{105}                                          & 105                                                      &33.8         &46.5     &**   \\
30 & AALS  \cite{wang2021knowledge}                & \multicolumn{1}{c}{105}                                                         & \multicolumn{1}{c}{105}               & \multicolumn{1}{c}{105}                                          & 105                                                      &33.8         &47.2      &**  \\
31 & OXNet  \cite{luo2021oxnet}                     & \multicolumn{1}{c}{105}                                                         & \multicolumn{1}{c}{105}               & \multicolumn{1}{c}{105}                                          & 105                                                      &34.4        &47.2   &*      \\
32 & UT  \cite{liu2021unbiased}                 & \multicolumn{1}{c}{105}                                                         & \multicolumn{1}{c}{105}               & \multicolumn{1}{c}{105}                                          & 105                                                      &34.3         &47.5     &*   \\
33 & ORF-Net \cite{chai2022orf}                  & \multicolumn{1}{c}{105}                                                         & \multicolumn{1}{c}{105}             &\multicolumn{1}{c}{105}                                             & 105                                                      &34.2         & 47.5  &** \\
34 & ORF-Netv2                 & \multicolumn{1}{c}{105}                                                         & \multicolumn{1}{c}{105}             &\multicolumn{1}{c}{105}                                             & 105                                                      &\textbf{34.7}         & \textbf{48.1}    &-     \\ \hline
\end{tabular}
\label{tab: SOTA RibFrac dataset}
\end{table}

\subsection{Comparison with the State-of-the-art}
We used the one-stage object detection model FCOS \cite{tian2020fcos} as our baseline model, which could be trained with box-labeled data.
To the best of our knowledge, few studies have been proposed to simultaneously leverage the box-labeled data, mask-labeled data, dot-labeled data, and unlabeled data for object detection.
Therefore, we compared ORF-Netv2 with SOTA semi-supervised object detection as well as the variants of these methods.
Specifically, we included: 1) STAC \cite{sohn2020simple}, which deployed highly confident pseudo labels from unlabeled images and trains the model with a strong augment strategy; 2) AALS \cite{wang2021knowledge}, a teacher-student framework with an adaptive asymmetric label sharpening algorithm; 3) Unbiased Teacher \cite{liu2021unbiased}, which leveraged the focal loss based on the teacher-student framework, and 4) ORFNet \cite{chai2022orf}, an omni-supervised framework with a multi-branch omni-supervised head proposed in our previous work.
We also modify these methods to enable them to be compatible with different annotations.
For the mask-labeled data, we generated the bounding boxes from the masks.
For the dot-labeled data, we computed the loss corresponding to only the positive dots and ignored the unlabeled samples.

\subsubsection{Results on RibFrac}
A quantitative comparison is reported in Table \ref{tab: SOTA RibFrac dataset}, where we changed the mixture of different types of data.
Additional mask-labeled data brought the greatest improvement compared with FCOS (2.9\% on mAP and 4.8\% on AP50, comparing rows 1 and 3), followed by dot-labeled data (1.1\% on mAP and 2.9\% on AP50, comparing rows 1 and 6), and finally unlabeled data (0.3\% on mAP and 0.9\% on AP50, comparing rows 1 and 10). 
When different types of data were combined, ORF-Netv2 achieved consistent improvement over other competitive methods, showing the superiority of utilizing as much supervision as possible.
For example, when combining box-labeled data, dot-labeled data, and unlabeled data for training, ORF-Netv2 improves 0.4\% mAP and 1.6\% AP compared with UT \cite{liu2021unbiased} (rows 12 and 14).
Moreover, when using box-labeled data, mask-labeled data, and unlabeled data to train the model, ORF-Netv2 improves 0.9\% mAP and 0.4\% AP compared with UT \cite{liu2021unbiased} (rows 16 and 18).
When all the different types of data were used, ORF-Netv2 achieved 34.7\% mAP and 48.1\% AP, which are the best performance among all SOTA methods.

\begin{figure*}[!t]
\begin{center}
	\includegraphics[width=1. \linewidth]{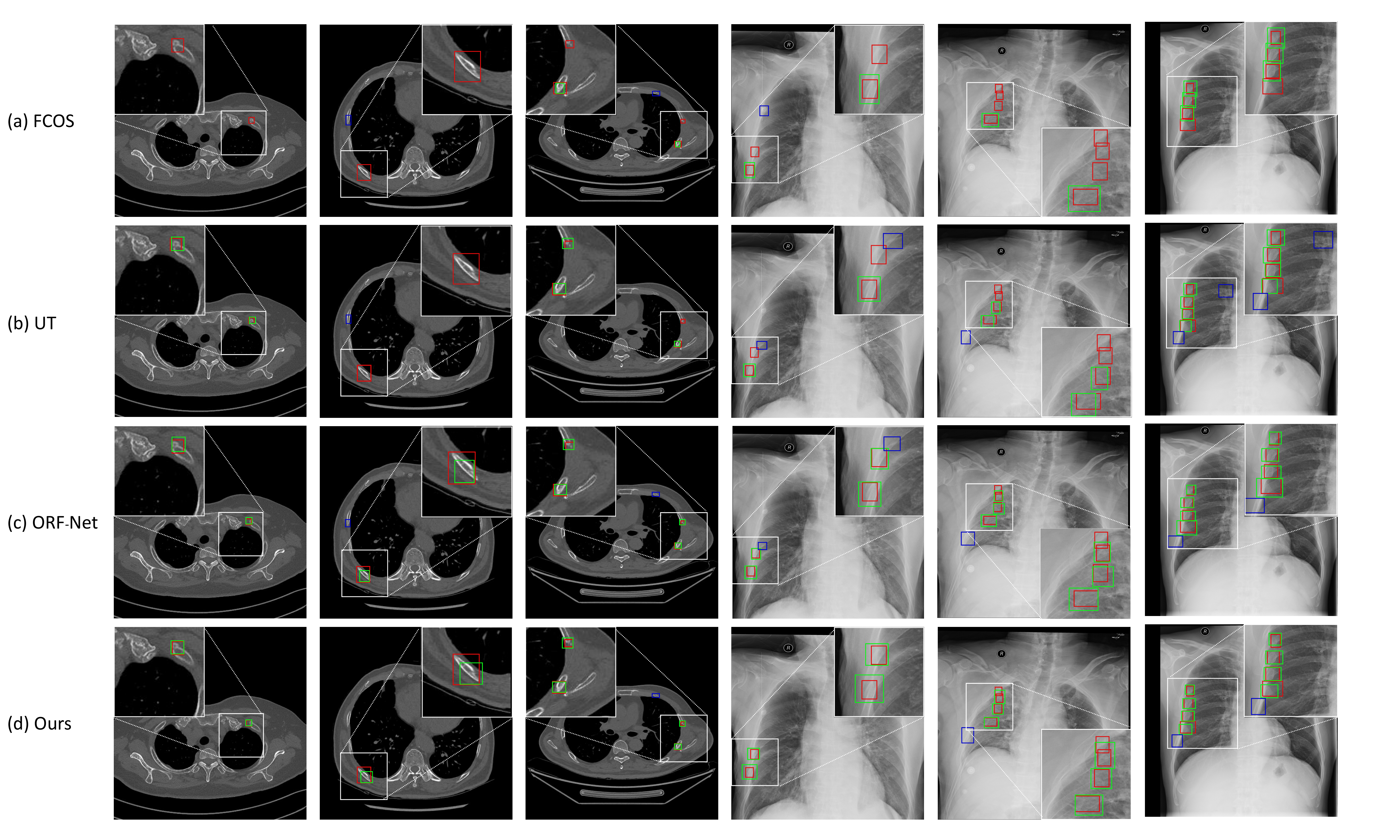}
\end{center}
\caption{Qualitative comparisons of the FCOS \cite{tian2020fcos}, UT \cite{liu2021unbiased}, ORF-Net \cite{chai2022orf}, and our proposed method on RibFrac and XRF. Ground truth, true positives, and false positives are annotated in \textcolor{red}{red boxes}, \textcolor{green}{green boxes}, and \textcolor{blue}{blue boxes}, respectively.}
\label{fig:VIS}
\end{figure*}

We further visualized a qualitative comparison in Fig. \ref{fig:VIS}.
It can also be observed that our proposed ORF-Netv2 detected the rib fractures from the CT images more correctly than other compared methods.

\begin{table}[t] \caption {Comparison with the SOTA on CRF. *: p-value $<$ 0.05; **: p-value $<$ 0.01.}
\color{black}
\centering
\begin{tabular}{ccccccc}
\hline
\multirow{2}{*}{Method} & \multicolumn{3}{c}{\#scans used}                                                                                                                                                                                 &\multicolumn{2}{c}{Metrics}  & \multirow{2}{*}{p-value} \\ \cline{2-6} 
                        & \multicolumn{1}{c}{\begin{tabular}[c]{@{}c@{}}$D_b$ \end{tabular}} & \multicolumn{1}{c}{\begin{tabular}[c]{@{}c@{}}$D_d$\end{tabular}} & \begin{tabular}[c]{@{}c@{}}$D_u$\end{tabular} & mAP          & AP50           \\ \hline
FCOS \cite{tian2020fcos}                    & \multicolumn{1}{c}{224}                                                         & \multicolumn{1}{c}{0}                                                           & 0                                                         & 39.9         & 53.7       & -  \\ \hline
FCOS \cite{tian2020fcos}                     & \multicolumn{1}{c}{224}                                                         & \multicolumn{1}{c}{450}                                                         & 0                                                         & 41.3         & 54.4       &**   \\
ORF-Net \cite{chai2022orf}                  & \multicolumn{1}{c}{224}                                                         & \multicolumn{1}{c}{450}                                                         & 0                                                         & 42.3         & 56.3       &**   \\
ORF-Netv2                  & \multicolumn{1}{c}{224}                                                         & \multicolumn{1}{c}{450}                                                         & 0                                                         & \textbf{42.6}         & \textbf{57.0}       & - \\

\hline
STAC \cite{sohn2020simple}                    & \multicolumn{1}{c}{224}                                                         & \multicolumn{1}{c}{450}                                                         & 1104                                                      & 40.0         & 56.1        &** \\
UT \cite{liu2021unbiased}                 & \multicolumn{1}{c}{224}                                                         & \multicolumn{1}{c}{450}                                                         & 1104                                                      & 42.6         & 56.3        &** \\ 
$\Pi$ Model \cite{laine2016temporal}                 & \multicolumn{1}{c}{224}                                                         & \multicolumn{1}{c}{450}                                                         & 1104                                                      & 42.9         & 56.3        &** \\ 
OXNet \cite{luo2021oxnet}                 & \multicolumn{1}{c}{224}                                                         & \multicolumn{1}{c}{450}                                                         & 1104                                                      & 42.9         & 56.5        &** \\ 
AALS \cite{wang2021knowledge}                  & \multicolumn{1}{c}{224}                                                         & \multicolumn{1}{c}{450}                                                         & 1104                                                      & 43.4        & 57.2       &**  \\ 
ORF-Net \cite{chai2022orf}     & \multicolumn{1}{c}{224}                                                         & \multicolumn{1}{c}{450}                                                         & 1104                                                      & 44.3        & 59.1       &*  \\
ORF-Netv2     & \multicolumn{1}{c}{224}                                                         & \multicolumn{1}{c}{450}                                                         & 1104                                                      & \textbf{44.7}        & \textbf{59.7}     &-  \\
\hline
\end{tabular}
\label{tab: SOTA CRF dataset}
\end{table}

\subsubsection{Results on CRF}
We also conducted experiments on the CRF dataset by comparing our method with previous works under different settings. 
With results in Table \ref{tab: SOTA CRF dataset}, we obtain the following observations.
First, FCOS \cite{tian2020fcos} can achieve an improvement of 1.4\% and 0.7\% on mAP and AP by simply learning from the labeled points on the dot-labeled data. 
Meanwhile, the proposed ORF-Netv2 achieves improvements on both mAP and AP50 (2.7\%, 3.3\%) compared with FCOS \cite{tian2020fcos}, which demonstrated the effectiveness of our proposed method on leveraging the dot-labeled data. 
Second, when incorporating unlabeled data, all the label-efficient learning methods improve over the supervised baseline, showing the effectiveness of these models in utilizing the unlabeled data. 
Finally, our proposed ORF-Netv2 outperforms all other models with at least 0.4\% in mAP, and 0.6\% in AP, demonstrating the effectiveness of omni-supervised learning in utilizing as much supervision as possible for rib fracture detection.

\begin{table}[t] \caption {Comparison with the SOTA on XRF. *: p-value $<$ 0.05; **: p-value $<$ 0.01.}
\color{black}
\centering
\begin{tabular}{ccccccc}
\hline
\multirow{2}{*}{Method} & \multicolumn{3}{c}{\#scans used}                                                                                                                                                                                 &\multicolumn{2}{c}{Metrics} & \multirow{2}{*}{p-value} \\ \cline{2-6} 
                        & \multicolumn{1}{c}{\begin{tabular}[c]{@{}c@{}}$D_b$ \end{tabular}} & \multicolumn{1}{c}{\begin{tabular}[c]{@{}c@{}}$D_d$\end{tabular}} & \begin{tabular}[c]{@{}c@{}}$D_u$\end{tabular} & mAP          & AP50           \\ \hline
FCOS \cite{tian2020fcos}                    & \multicolumn{1}{c}{2154}                                                         & \multicolumn{1}{c}{0}                                                           & 0                                                         & 14.4         & 24.2        &- \\ \hline
FCOS \cite{tian2020fcos}                     & \multicolumn{1}{c}{2154}                                                         & \multicolumn{1}{c}{2154}                                                         & 0                                                         & 15.7         & 26.3     &-   \\
ORF-Net \cite{chai2022orf}                  & \multicolumn{1}{c}{2154}                                                         & \multicolumn{1}{c}{2154}                                                         & 0                                                         & 17.4         & 28.5        &** \\
ORF-Netv2                  & \multicolumn{1}{c}{2154}                                                         & \multicolumn{1}{c}{2154}                                                         & 0                                                         & \textbf{19.0}         & \textbf{31.7}     &-   \\

\hline
$\Pi$ Model \cite{laine2016temporal}                   & \multicolumn{1}{c}{2154}                                                         & \multicolumn{1}{c}{2154}                                                         & 2154                                                      & 17.9         & 30.8      &**   \\
STAC \cite{sohn2020simple}                    & \multicolumn{1}{c}{2154}                                                         & \multicolumn{1}{c}{2154}                                                         & 2154                                                      & 18.3         & 29.1        &** \\
AALS \cite{wang2021knowledge}                  & \multicolumn{1}{c}{2154}                                                         & \multicolumn{1}{c}{2154}                                                         & 2154                                                      & 18.9        &  31.0     &**  \\ 
UT \cite{liu2021unbiased}                 & \multicolumn{1}{c}{2154}                                                         & \multicolumn{1}{c}{2154}                                                         & 2154                                                      & 19.2         & 31.5        &* \\ 
OXNet \cite{luo2021oxnet}                 & \multicolumn{1}{c}{2154}                                                         & \multicolumn{1}{c}{2154}                                                         & 2154                                                      & 19.1         & 32.0       &*  \\ 
ORF-Net \cite{chai2022orf}     & \multicolumn{1}{c}{2154}                                                         & \multicolumn{1}{c}{2154}                                                         & 2154                                                      & 19.2        & 31.9        &* \\
ORF-Netv2     & \multicolumn{1}{c}{2154}                                                         & \multicolumn{1}{c}{2154}                                                         &2154                                                      & \textbf{19.4}        & \textbf{32.8}       &-  \\
\hline
\end{tabular}
\label{tab: SOTA XRF dataset}
\end{table}

\subsubsection{Results on XRF} To verify the scalability of ORF-Netv2, we applied it on the chest X-rays with experiments on the XRF dataset.
The quantitative experimental results are reported in the table \ref{tab: SOTA XRF dataset}. 
When combing box-labeled data and dot-labeled data for training, ORF-Netv2 achieved 19.0\% mAP and 31.7\% AP50, with improvements of 1.6\% mAP and 3.2\% AP50 compared with ORF-Net. 
When all types of data were utilized, ORF-Netv2 consistently achieved the best performance on both mAP (19.4\%) and AP50 (32.8\%) compared to all other methods, demonstrating the benefits and flexibility of the model in taking advantage of all kinds of supervision.

We also compared ORF-Netv2 with other methods qualitatively with visualization shown in Fig. \ref{fig:VIS}, where our proposed model accurately detected multiple fractures on the chest X-ray images.

\subsubsection{Computational Cost}
\textcolor{black}{
Training ORF-Netv2 with different numbers of branches requires different time.
It took about 14.6 h, 16.7 h, 22.5 h, and 26.3 h to train ORF-Netv2 with a single branch (which is similar to the time used by FCOS), two branches, three branches, and four branches, respectively.
For inference, ORF-Netv2 has a larger model size compared with that of FCOS (150 MB vs. 123 MB) as the former added multiple classification branches to learn from different types of supervision.
ORF-Netv2 also takes longer inference time compared to FCOS (0.15 s/slice vs. 0.06 s/slice).
All other compared methods, i.e., $\Pi$ model, STAC, AALS, OXNet, and UT, share similar inference structures and time with FCOS.
However, as discussed in Table \ref{tab: cls branches}, different branches will converge to similar performance, and one may also consider using a single branch for lighter and faster inference with slight sacrifice on the performance.
}

\subsection{Budget-aware Omni-supervised Detection}

Annotating medical images is labor-tedious and expertise-depending.
\textcolor{black}{In the previous experiments, we have shown that ORF-Netv2 consistently outperforms other competitive methods under various combinations of annotations.
In this section, to help practical model development, we further explore which labeling policy will bring greater benefits to ORF-Netv2 under a limited budget.
}
We started by evaluating the time of conducting different annotations on a subset of RibFrac. 
The average time to generate dot annotations, bounding boxes, and masks to mark rib fractures slice by slice on a chest CT scan was approximately 228 seconds, 305 seconds, and 629 seconds, respectively.
We then studied different labeling policies under a fixed labeling budget of 66,000 seconds.
Specifically, four policies were taken into consideration: (1) STRONG-B: all the budget is used to annotate bounding boxes; (2) STRONG-M: all the budget are used to annotate masks; (3) EQUAL: using one-third of the budget for each type of annotation; (4) EQUAL-NUM: labeling same amount of data for each type.

\begin{table}[ht] \caption {Budget-aware omni-supervised rib fracture detection on RibFrac.}
\centering
\begin{tabular}{ccccccc}
\hline
\multirow{2}{*}{Policy} & \multicolumn{4}{c}{\#scans used}                                                                        &\multicolumn{2}{c}{Metrics} \\ \cline{2-7} 
                        & \multicolumn{1}{c}{\begin{tabular}[c]{@{}c@{}}$D_b$\end{tabular}}
                        & \multicolumn{1}{c}{\begin{tabular}[c]{@{}c@{}}$D_m$\end{tabular}}& \multicolumn{1}{c}{\begin{tabular}[c]{@{}c@{}}$D_d$\end{tabular}} & \begin{tabular}[c]{@{}c@{}}$D_u$\end{tabular} & mAP          & AP50           \\ \hline

STRONG-B                & \multicolumn{1}{c}{217}                                                         & \multicolumn{1}{c}{0}             &\multicolumn{1}{c}{0}                                             & 203                                                      &\textbf{33.9}        & \textbf{46.3}        \\ \hline
STRONG-M                & \multicolumn{1}{c}{0}                                                         & \multicolumn{1}{c}{105}             &\multicolumn{1}{c}{0}                                             & 315                                                      &31.9        &43.8        \\ \hline
EQUAL               & \multicolumn{1}{c}{72}                                                         & \multicolumn{1}{c}{97}             &\multicolumn{1}{c}{35}                                             & 216                                                      &31.9         & 44.1        \\ \hline

EQUAL-NUM                & \multicolumn{1}{c}{57}                                                         & \multicolumn{1}{c}{57}             &\multicolumn{1}{c}{57}                                             & 249                                                      &32.3         & 45.0         \\ \hline
\end{tabular}
\label{tab: budget}
\end{table}

As reported in Table \ref{tab: budget}, under a limited labeling budget, we found that policy STRONG-B achieved a large improvement (2.\% mAP and 2.5\% AP 50) than policy STRONG-M, which indicated that the cost performance of labeling box was much higher than labeling mask in the task of rib fracture detection. 
This observation is in line with the clinical insight that radiologists care more about the detection rate of rib fractures than delineating the ambiguous boundaries of the lesions.
Moreover, we also observed that the policy EQUAL-NUM performed better than the policy EQUAL, which suggested that making efforts to generate the same amount of different annotations leads to more improvement to our model.
Despite that the STRONG-B policy achieved the best performance here, the significance of omni-supervised learning is to use as much supervision as possible to improve the performance.
Therefore, when the budget allows for more labels with different annotation forms, our model can continuously exploit various types of data to achieve better performance.

\section{Discussion}\label{disscussion}
Rib fracture detection is an important task in clinical diagnosis, and accurate identification as well as localization of rib fractures can significantly improve the outcome of patients with thoracic trauma. 
Although recent deep learning-based fracture detection approaches have shown remarkable progress, they mostly relied on supervised training with a large number of fine-grained annotations, which posed a huge burden on data acquisition and labeling. 
In clinical practice, there are usually multiple types of data with different annotation forms, such as mask-labeled data, box-labeled data, dot-labeled data, and unlabeled data that we have explored in this study.
\textcolor{black}{
Neither semi-supervised learning nor weakly-supervised learning could sufficiently address the challenge of a growing variety of annotation types.
Nevertheless, only a dearth of omni-supervised works have attempted to exploit all these available supervision to improve detection performance.
Therefore, to our knowledge, we proposed the first omni-supervised rib fracture detection framework, ORF-Netv2, to exploit multiple granularities of annotations for label-efficient and annotation-friendly medical image analysis.
}

\begin{figure}[!t]
\begin{center}
	\includegraphics[width=1. \linewidth]{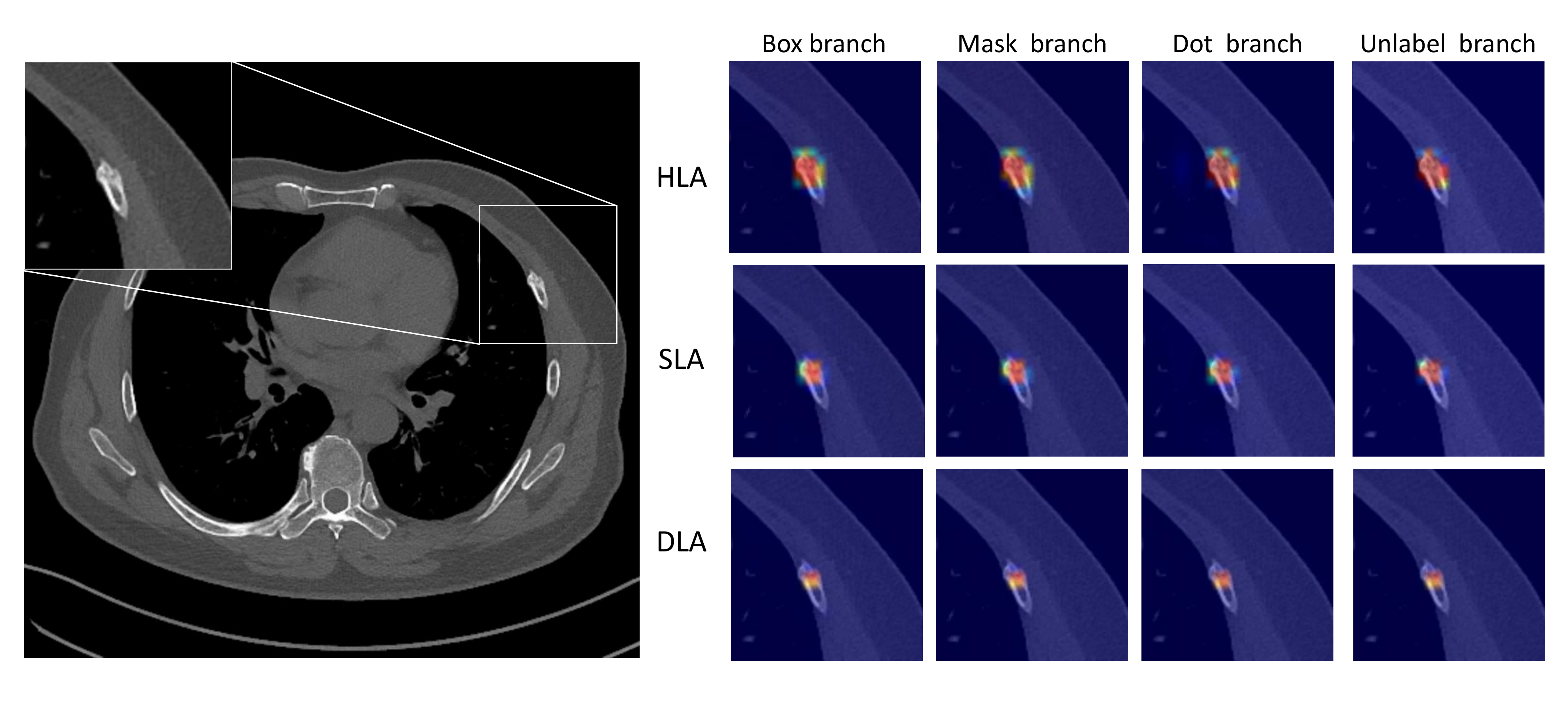}
\end{center}
\caption{\textcolor{black}{Visualization of the prediction probability maps with different label assignment strategies.}}
\label{Viz_LA_comparison}
\end{figure}

Moreover, different from the existing omni-supervised object detection methods which were mostly based on generating pseudo labels with a teacher model \cite{ren2020ufo,luo2021oxnet,wang2022omni}, we proposed the co-training-guided label assignment strategies based on a multi-branch co-training scheme.
Particularly, the most fine-grained annotations (i.e., the polygon masks) may not clearly define the pixel samples of the target lesion.
Our proposed method tackles the challenge of label assignment for fully-labeled, weakly-labeled, and unlabeled data in a unified manner with great flexibility and robustness.
\textcolor{black}{We show with Fig. \ref{Viz_LA_comparison} that ORF-Netv2 could learn clearer boundaries between the positive and negative pixels. This indicates that ORF-Netv2 could focus more on important samples and reduce the contribution of low-quality samples with the proposed learnable label assignment strategy.}
\textcolor{black}{In this way, we eliminate the reliance on pseudo labels for OSL-based detection and also shed light on extending the dynamic label assignment to situations where fully-labeled data are insufficient.}
Extensive experiments demonstrated the effectiveness of ORF-Netv2 with consistent improvement over other competitive methods.

\textcolor{black}{We took FCOS as the base model for all experiments because its anchor-free structure offers more flexibility in designing label assignment strategies for the tiny and ambiguous rib fractures.
FCOS has also been demonstrated as a powerful object detector by many other recent studies \cite{liu2023decoupled, li2023one, he2022research, jiao2023attention}.
To further verify the selection of our backbone, we compared FCOS with other supervised object detectors (Faster r-cnn \cite{ren2015faster}, RetinaNet \cite{lin2017focal}, and Deformable DETR \cite{zhu2021deformable}) on the box-labeled data in RibFrac dataset. 
By the results reported in Table \ref{tab:backbone}, FCOS outperformed other compared approaches, showing more effectiveness on detecting rib fractures.
}

\textcolor{black}{We noticed that the mAP performance of the models varied on the two CT datasets, which is mainly due to the annotation differences between the datasets. Specifically, in the RibFrac dataset, the size of each rib fracture box fluctuated widely with sizes ranging from 27 to 3822 pixels.
In the CRF dataset, the size of each fracture box does not vary greatly with sizes ranging from 228 to 1044 pixels.
To verify this, we further tested ORF-Netv2 on RibFrac for the lesions with sizes ranging from 228 to 1044 pixels, and the mAP was much closer to that on CRF (46.0\% vs 44.7\%).}

A potential limitation of the current study is that we only focused on detecting rib fractures, and the performance of detecting other lesions or objects remains to be further explored in the future.
Nevertheless, the former presented extensive experiments demonstrated the generality of ORF-Netv2, showing a promising method that can be easily extended to other object detection tasks.

\begin{table}[t] \caption {\textcolor{black}{Comparison of different supervised object detectors on the box-labeled data from RibFrac.}}
\color{black}
\centering
\begin{tabular}{cccc}
\hline
\multirow{2}{*}{Method} & \multicolumn{1}{c}{\#scans used}    &\multicolumn{2}{c}{Metrics} \\ \cline{2-4} &$D_b$ &mAP          & AP50           \\ \hline
FCOS \cite{tian2020fcos}               & \multicolumn{1}{c}{105}    &\textbf{30.7}         &\textbf{42.5}         \\ \hline
Faster RCNN \cite{ren2015faster}                & \multicolumn{1}{c}{105}           &29.5        & 40.9 \\ \hline    
RetinaNet \cite{lin2017focal}               & \multicolumn{1}{c}{105}      &28.7         & 40.1       \\ \hline
Deformable DETR  \cite{zhu2021deformable}             & \multicolumn{1}{c}{105}         &28.5        & 38.9 \\ \hline

\end{tabular}

\label{tab:backbone}
\end{table}

\section{Conclusion}\label{conclusion}
In this paper, we explore and verify the effectiveness of omni-supervised learning in rib fracture detection by the proposed ORF-Netv2, a unified framework that utilizes as much available supervision as possible.
To enable omni-supervised detection, we design an omni-supervised detection network with a novel co-training-guided dynamic label assignment strategy to learn from the diversely annotated data in a holistic manner.
Extensive experiments on three typical rib fracture detection chest radiology datasets demonstrate the effectiveness of our method in utilizing the various granularities of supervision.
Moreover, ORF-Netv2 is flexible and general, which can be easily extended to other tasks of object detection.

\bibliographystyle{IEEEtran}
\bibliography{reference}

\end{document}